# A Dynamic Robotic Actuator with Variable Physical Stiffness and Damping

Manuel Aiple[1], Wouter Gregoor[2], André Schiele[1]

*Delft University of Technology, Mekelweg 2, 2628 CN Delft, Netherlands*

**Abstract**

This study is part of research aiming at increasing the range of dynamic tasks for teleoperated field robotics in order to allow operators to use the full range of human motions without being limited by the dynamics of the robotic manipulator. A new variable impedance actuator (VIA) was designed, capable of reproducing motions through teleoperation from precise positioning tasks to highly dynamic tasks. The design requirements based on previous human user studies were a stiffness changing time of 50 ms, a peak output velocity of 20 rad/s and variable damping allowing to suppress undesired oscillations. This is a unique combination of features that was not met by other VIAs. The new design has three motors in parallel configuration: two responsible for changing the VIA's neutral position and effective stiffness through a sliding pivot point lever mechanism, and the third acting as variable damper. A prototype was built and its performance measured with an effective stiffness changing time of 50 ms to 120 ms for small to large stiffness steps, nominal output velocity of 16 rad/s and a variable damper with a damping torque from 0 N m to 3 N m. Its effective stiffness range is 0.2 N m/rad to 313 N m/rad. This concludes that the new actuator is particularly suitable for highly dynamic tasks. At the same time, the new actuator is also very versatile, making it especially interesting for teleoperation and human-robot collaboration.

## 1. Introduction

Today, teleoperation is mostly used for performing slow motions for precision tasks like picking and placing objects or guiding tools. Therefore, teleoperated robots are typically optimized for precision instead of dynamics. In field robotics however, like future space exploration or inspection of disaster sites, flexibility and improvisation can be very advantageous, *e.g.*, to clear the path for a robot deployed in the Fukushima Daiichi nuclear disaster [1]. Humans are able to perform a wide range of motions, including very dynamic ones like throwing and hammering, allowing them to cope with unexpected situations. Still, the current teleoperation systems are not capable of reproducing these, which can be a limiting factor for telerobotics in unstructured environments.

In a previous study we could show with a very simple teleoperated series elastic actuator (SEA) that humans are able to exploit the resonance of a flexible tool device in order to reach a higher velocity at the hammer head than the actuator's motor velocity when hammering [2]. Nevertheless, the SEA used in that study was not suitable for performing precise positioning tasks as the high compliance made it hard to position in the presence of external forces and its lack of damping meant that it oscillated for a long time after moving. But a truly useful actuator for teleoperation in the field needs to allow to perform positioning tasks as well as dynamic tasks, suggesting the use of variable impedance actuators (VIAs). VIAs are currently the best candidate technology for building a telemanipulated robot appropriate for wide scope teleoperation as their compliance can be mechanically adapted to the needs of the task. They can be used as rigid actuators for positioning tasks or as soft actuators for highly dynamic tasks, as has been shown for various applications [3, 4, 5, 6, 7, 8, 9, 10, 11, 12].

A suitable VIA to continue our research should match our previous findings concerning the frequency and velocity of teleoperated hammering [2], but also provide variable stiffness and damping to increase precision and reduce oscillations compared to an SEA. Furthermore, Garabini *et al.* concluded that the peak output velocity achieved when hammering with a VIA can be increased by 30 % if the stiffness is adapted during the hammering motion compared to keeping the stiffness constant [13]. Thus, the stiffness changing mechanism of the VIA should be fast enough to allow stiffness changes during the hammering motion. No existing design was found that could match all requirements. Goal of this paper is therefore to present the design of a dynamic robotic actuator (Dyrac) meeting the mentioned requirements, resulting in a unique actuator prototype for dynamic teleoperation research. While Dyrac is primarily designed for this research, it should also be useful as a generic VIA, and hence was designed to have an output torque and stiffness range comparable to other VIAs.

---

★Corresponding author M. Aiple: `m.aiple@tudelft.nl`

[1]M. Aiple and A. Schiele are with the BioMechanical Engineering Department, Faculty of Mechanical, Maritime and Materials Engineering, Delft University of Technology, Netherlands.

[2]W. Gregoor is with the Electronic and Mechanical Support Division, Delft University of Technology, Netherlands.

## 2. Design

### 2.1. Considerations

In the following, a subset of VIA design options is summarized that were considered in the Dyrac design, disregarding options that rely on modifying the stiffness of the elastic element itself. A broader and more systematic classification of VIA designs has for example been done by Vanderborght *et al.* [11] and Wolf *et al.* [12], which is reflected in the following as far as relevant for the Dyrac design.

The joint position of VIAs depends on the *neutral position*, which is the joint position in absence of external forces, the *effective stiffness* of the VIA, which is the stiffness of the elastic element multiplied by the stiffness variation factor, and the external forces acting on the joint output that cause a deflection of the joint. All VIAs need at least two motors to change the effective stiffness and the neutral position of the actuator independently. These can be configured in parallel, with both motors mounted to the chassis, or in series, with the second motor attached to the output axis of the first motor. Typically, both motors are equally strong for actuators in parallel configuration, whereas in series configuration the second motor is typically smaller to reduce the mass that needs to be moved by the first motor. In parallel configuration, the two motors typically work in opposite directions for changing the stiffness (as antagonists) and in the same direction for changing the neutral position, *e.g.*, the DLR BAVS [14]. In series configuration, only the first motor influences the neutral position, whereas the second motor only influences the stiffness, *e.g.*, the IIT AwAS-II [15]. Designs using the parallel configuration typically have quicker stiffness changing times, whereas designs using the series configuration typically are more compact and lighter as only one powerful motor is required, while the second one can be relatively small.

Besides the choice of the motor configuration, the mechanism for changing the stiffness variation factor influences the performance of the VIA significantly. Cam-roller mechanisms couple the joint deflection to the spring deflection through a roller moving on a non-linearly shaped cam profile, *e.g.*, the DLR FSJ [16]. The cam profile is shaped in such a way that as the deflection increases the slope of the cam-profile increases as well, and thereby the effective stiffness increases. The stiffness at zero deflection can be varied by changing the initial position of the roller on the cam profile, for example by having two cam profiles which can be moved relative to each other. Cam-roller mechanisms can be made very compact, as the slope of the cam profile is relevant, but not the absolute height of the profile. Cam-roller mechanisms always have a non-linear torque over deflection curve, as this is required to be able to change the stiffness, which might be desirable or not, depending on the application. Also, typically the stiffness range that can be obtained is relatively small, as it is limited by the dimensions of the cam profile.

Lever mechanisms use a lever arm to reduce or increase the spring deflection compared to the joint deflection, *e.g.*, the vsaUT-II [17]. For varying the stiffness, the lever arm length can be modified, or the application point of the force on the lever arm can be moved, or the pivot point of the lever. Lever mechanisms can have a linear torque over deflection curve, as the stiffness variation factor is not influenced by the deflection. When changing the stiffness by moving the lever pivot point, the stiffness can be varied from zero to infinity (in practice this is limited by the material stiffness). Furthermore, the stiffness can be changed while keeping the potential energy constant that is stored in the spring. Often lever mechanism based designs take more space to accommodate for the lever.

Different physical principles can be and have been used to implement variable physical damping for robotic actuators, *e.g.*, using friction brakes [18], Eddy currents [19], magnetorheological fluids [20], or an electric motor in direct drive configuration. Friction brakes are compact and can be disengaged fully to present zero damping torque, however the torque is difficult to control precisely due to the complex friction processes. Eddy current based dampers require powerful electromagnets to induce the braking current into the rotor, but the velocity-dependency of the torque is part of the physical principle, leading to a behavior very similar to the desired viscous damping behavior. Dampers based on magnetorheological fluids allow to render viscous damping very accurately, but they are complex to design and implement, requiring complex multiphysics finite element method (FEM) analysis and a mechanical design preventing leaking of the fluid. Electric motors are easy to integrate in an actuator and to control in torque mode for rendering viscous damping behavior. However, they are relatively big and the damping torque is limited, meaning that the maximum damping factor that can be rendered decreases with increasing speed. Also, they can lead to unstable behavior of the actuator, thus requiring fine tuning of the torque controller.

Table 1: Comparison of selected existing VSA designs to the Dyrac requirements

| Design | Requirements (Prototype Characteristics) | BAVS [14] | FSJ [16] | MACCEPA [21] | AwAS-II [15] | vsaUT-II [17] | VSA-Cube with Damper Module [22, 23] |
|---|---|---|---|---|---|---|---|
| Motor Configuration | — (Parallel) | Parallel | Serial | Serial | Serial | Parallel | Parallel |
| Stiffness Variation Mechanism | — (Lever pivot point) | Cam-roller | Cam-roller | Lever arm length | Lever pivot point | Lever pivot point | Agonistic-antagonistic |
| Nominal velocity | 20 rad/s (16 rad/s) | 12.6 rad/s | 8.5 rad/s | 5.8 rad/s | 10.2 rad/s | 2.2 rad/s | 3 rad/s |
| Maximum torque | — (7 N m) | 8 N m | 67 N m | 70 N m | 80 N m | 60 N m | 3 N m |
| Stiffness changing time | 50 ms (50 ms to 120 ms) | 14 ms | 330 ms | 2600 ms | 800 ms | 500 ms | 180 ms to 320 ms |
| Minimum stiffness | — (0.2 N m/rad) | 3.9 N m/rad | 52.4 N m/rad | 5 N m/rad | 0 N m/rad | 0 N m/rad | 3 N m/rad |
| Maximum stiffness | — (313 N m/rad) | 146.6 N m/rad | 826 N m/rad | 110 N m/rad | ∞ | ∞ | 14 N m/rad |
| Variable Damping | Yes (damping torque 3 N m) | No | No | No | No | No | Yes |

*2.2. Requirements*

Our previous research showed that humans execute hammering motions at a frequency around 5 Hz and a peak velocity of 20 rad/s [2]. Furthermore, a stiffness changing time under 50 ms is required to perform the hard-soft-hard-soft cycle described in [13] within the measured typical hammering period of 200 ms. We considered it highly advisable to add variable physical damping as requirement to the actuator, as the importance of variable physical damping for oscillation suppression has been highlighted for example by Laffranchi *et al.* [24]. Table 1 summarizes the design requirements and compares them to the performance of selected existing VIA designs.

The primary design goal of this study was to match the requirements of table 1, and the secondary design goal was to develop a very versatile actuator, allowing to explore the parameter space velocity – stiffness – damping extensively. Existing actuator designs with similar performance as the requirements are among others: the Bidirectional Antagonistic Variable Stiffness joint (BAVS) [14], the Floating Spring Joint (FSJ) [16], the Mechanically Adjustable Compliance and Controllable Equilibrium Position Actuator (MACCEPA) [21], the Actuator with Adjustable Stiffness II (AwAS-II) [15], and the Variable Stiffness Actuator University Twente II (vsaUT-II) [17]. The BAVS [14] and the FAS [25] from DLR allow a stiffness change within the required time, but BAVS only reaches an angular velocity of 12.6 rad/s, and FAS is too small as it was designed to actuate a finger tendon, and neither has variable damping. Catalano *et al.* have implemented a variable damping module [23] to be combined with the VSA-Cube [22] to form a VIA with variable physical stiffness and damping similar to the requirements of this study, but too slow with a maximum velocity of 4.7 rad/s.

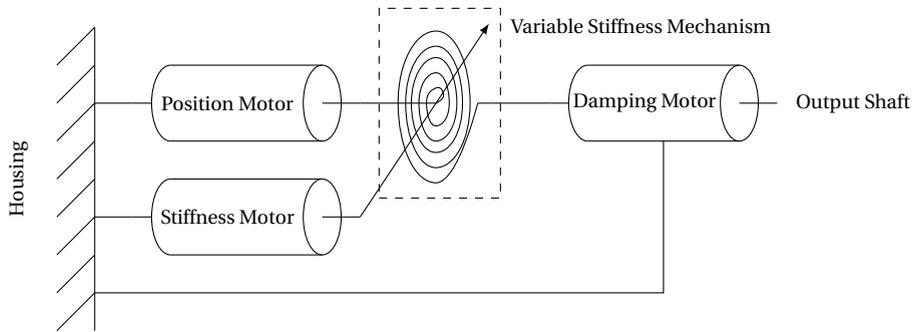

Figure 1: High level configuration of the actuator. The actuator has three motors: the position motor changes the neutral position of the compliant joint, the stiffness motor changes the stiffness, and the damping motor serves to implement variable damping. Position motor and stiffness motor are mounted in parallel to the housing, thus the stiffness motor has to move with the position motor to change the neutral position without changing the stiffness.

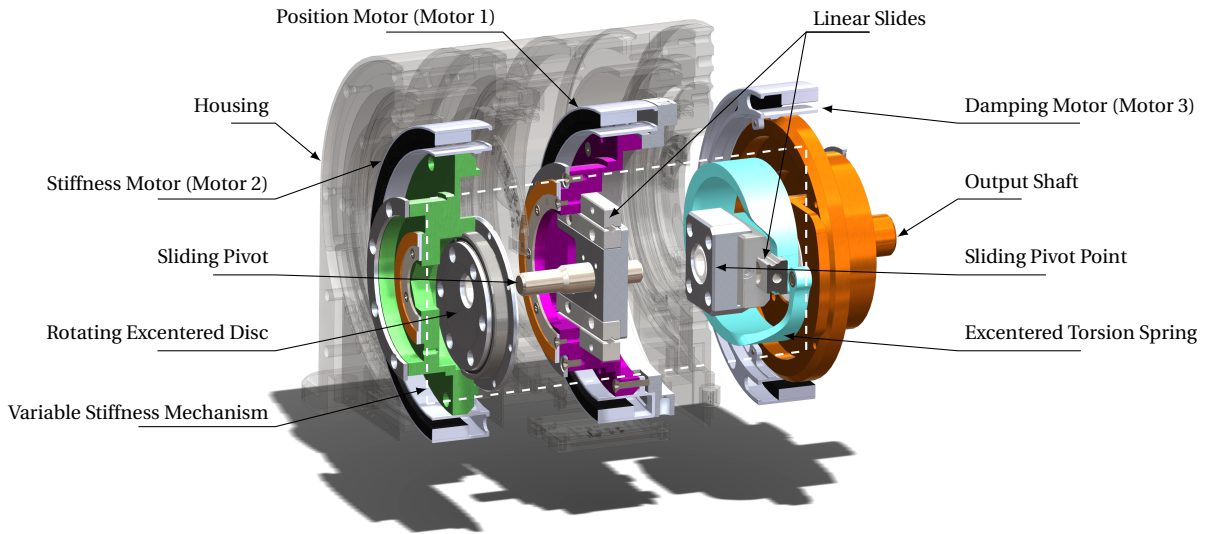

Figure 2: Exploded view of the Dyrac prototype CAD model. The stators of the motors are fixed to the housing (greyed out) and the rotors of the motors all rotate around a common axis with the output shaft. The position motor turns the linear slide holding the sliding pivot, which carries the pivot point on the spring with it when the pivot is out of center. Depending on the position of the pivot on the linear guide, the lever ratio for a force acting on the excentered torsion spring changes, thus changing the effective stiffness at the output shaft. The position of the pivot on the linear guide can be changed by rotating the stiffness motor with respect to the position motor through the crankshaft mechanism formed by the rotating excentered disc. The damping motor sits on the output shaft implementing a variable damping.

*2.3. Actuator Principle*

Based on the requirements of section 2.2 and the considerations of section 2.1 the architecture chosen for Dyrac is a parallel motor configuration design with sliding pivot point lever stiffness variation mechanism, similar to the vsaUT-II, with a variable damper based on an electric motor in direct drive configuration. This architecture was chosen as the parallel motor configuration adds less inertia on the motor changing the neutral position, helping to achieve fast acceleration of the end-effector. For the stiffness variation mechanism, a lever arm architecture with changing pivot point was chosen as it allows to change the stiffness over the full stiffness scale, thus making the actuator useful for a wide range of studies to explore the velocity – stiffness – damping parameter space. An electric motor was chosen to implement the damper as it is not only much simpler to realize with commercially available components, but also allows to actively influence the output torque for precise torque servoing, making the actuator even more versatile. Again, the design was made such that the damping motor acts in parallel with the other motors, mounted to the chassis. This design choice makes the overall architecture similar to the DM$^2$ approach in [26], which was shown to allow a recovery of performance lost by compliant actuators up to a certain extent [3]. However, here, the damping capability was the primary objective of this third motor.

Figure 1 shows the high level configuration of Dyrac and figure 2 shows an exploded view of the prototype CAD model. The actuator has three motors, a position motor for changing the neutral position, a stiffness motor for changing the effective stiffness, and a damping motor for variable damping. All motors and the output shaft are arranged on a common center axis in three stages.

The position motor is in the center, rotating a sliding table on which the pivot is fixed. When the pivot is excentered, it can excert a force in tangential direction to induce torque on the output shaft. This happens through the pivot point mounted to the torsion spring, connected to the output shaft. The effective stiffness of the actuator, *i.e.*, the stiffness as measured on the output shaft, depends on the distance of the pivot to the center axis, because the spring is not located at the center axis, but excentered, thus the lever from the pivot compared to the output shaft changes in opposite way to the lever from the pivot compared to the excentered torsion spring, when the pivot position changes (cf. Figure 3). Thus, for the same torque on the output shaft, the force exerted by the spring on the pivot is different, depending on the position of the pivot because of the lever effect between force and torque, which means that the effective stiffness can be changed by changing the pivot position.

The stiffness motor on the left in figure 2 is used to change the pivot distance from the center axis through a crankshaft mechanism formed by an excentered disc to which the pivot is connected through a bearing, whose rotation axis is also excentered relative to the excentered disc. Thus, by rotating the stiffness motor relative to the position motor, the pivot slides along the linear slide to a new distance from the center axis. The damping motor on the right in figure 2 is controlled in torque control mode and directly coupled to the output shaft, thereby allowing to implement variable damping.

*2.4. Kinematics*

Figure 3 shows the kinematics of the actuator viewed from the front with the center axis from left to right and figure 4 shows the name conventions and geometric relations as seen from the left with the center axis pointing out of the paper. The angular polar coordinate of the pivot point $P$ will be referred to as *pivot angle $\varphi$* and the radial polar coordinate as *pivot radius $r$*.

Three basic actions can be performed on the actuator: change of neutral position, change of stiffness, and deflection.

1. The neutral position is changed by equal rotation of motor 1 and 2. This causes a change of the pivot angle $\varphi$ while keeping the pivot radius $r$ constant.
2. The stiffness is changed by rotation of motor 2 relative to motor 1 (change of the angle $\varphi_d$) through the crankshaft mechanism formed by the excentered disc $\overline{BP}$ rotating relative to the link $\overline{OB}$ to motor 2. This causes $P$ to move along the two linear slides represented by the lines going through $\overline{OC}$ and through $\overline{PA}$. Notice that a movement of $P$ along $\overline{OC}$ also requires a change of either $\alpha$ or $\theta$ to maintain the triangle $OPA$, corresponding to a change of the output shaft position or of the spring deflection, *i.e.*, of the output torque. Changing the position of $P$ and thus the length of $r = \overline{OP}$ changes the effective stiffness of the spring by changing the length of the spring lever length $c = \overline{AP}$. Indeed, if $P$ is on one axis with the output axis going through $O$ for $r = 0$, the output axis can rotate freely around the pivot axis and the spring has no effect, which corresponds to an effective stiffness of zero. On the other hand, if $P$ is on one axis with the spring rotation axis going through *A* for $r = l$, then the output axis is locked to the motor 1 axis and the effective stiffness of the spring is infinite. For any other values of *r* between these two extremes, the effective stiffness has a certain finite value, with the special case of $r = l/2$, for which the effective stiffness equals the nominal stiffness of the spring.

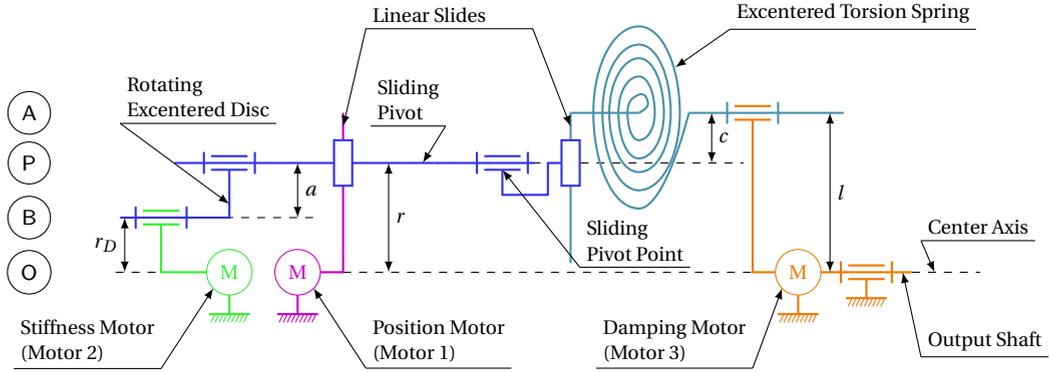

Figure 3: Front view of the kinematics of the actuator. Motor 1 sets the neutral position of the actuator. Moving motor 2 relative to motor 1 changes the position of the sliding pivot through the slider-crank mechanism realized with the excentered disc connected to motor 2 and the linear slide connected to motor 1. The change of the pivot position on the linear slide in turn changes the position of the sliding pivot point on the linear slide mounted on the excentered torsion spring, which changes the effective stiffness of the actuator, by changing the lever ratio between spring input (motor 1) and output (output shaft). A torque on the output shaft causes a deflection of the spring and a counter-force depending on the sliding pivot position. If the sliding pivot is aligned with the center axis, the actuator has zero stiffness as the output shaft can turn freely around the pivot point. If the sliding pivot is aligned with the spring axis, the actuator has infinite stiffness as the spring cannot be deflected. Generating a torque with motor 3 can be used to dampen oscillations on the output shaft. The dimensions $r_D$, $a$ and $l$ are design parameters, whereas $r$ and $c$ change during operation of the actuator. The positions notated A, P, B and O on the left refer to the projections of the points used in figure 4.

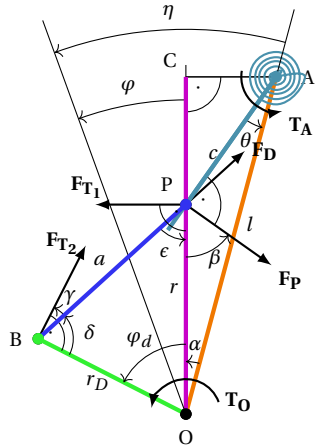

Figure 4: Geometric relations of the kinematics (view from the left compared to figure 3). The neutral position $\varphi$ is controlled by motor 1 (purple). The angular position $\eta$ of the output axis (orange) is the sum of neutral position $\varphi$ and the deflection angle $\alpha$. The triangle $OBP$ is defined by the lengths $a$ and $r_D$ and the angle $\varphi_d$. Changing $\varphi_d$ by moving motor 2 (green) relative to motor 1 will modify the pivot radius $r$. The triangle $OPA$ is defined by the lengths $r$ and $l$ and the angle $\alpha$. Changing $\alpha$ by deflecting the output axis (orange) compared to the neutral position will modify $\theta$, which is the spring deflection angle, and result in a counter-force from the spring deflection. The other measures in the figure are indicated to help obtain equations 4 to 10.

3. A torque $T_O$ on the output shaft while motor 1 and 2 maintain their positions causes a deflection of the spring by the angle $\theta$ and an output deflection angle $\alpha$. The torque $T_O$ required to cause a deflection of $\alpha$ is determined by the effective stiffness. Notice that this is only possible because *c* is not constant, but can vary when the spring is deflected thanks to the linear slide. On the other hand, this also means that the stiffness is not constant over the deflection, but decreases with increasing deflection.

Equations 1 to 16 describe the relationships to calculate the output torque $T_O$ (eq. 10) and the torques $T_1$ and $T_2$ of motor 1 and 2 (eq. 15 and 16) depending on the design parameters $r_D$, *a* and *l*, the spring stiffness *k*, the stiffness setting angle $\varphi_d$, and the deflection $\alpha$ for a static equilibrium of forces. Some helping variables $\beta$, $\gamma$, and $\delta$ can easily be obtained from the basic trigonometric relations:

$$\beta = \frac{\pi}{2} - \alpha - \theta, \tag{1}$$

$$\gamma = \frac{\pi}{2} - \delta, \tag{2}$$

and

$$\delta = \pi - \varphi_d - \epsilon. \tag{3}$$

From the law of sines, one obtains $\epsilon$ and *r*:

$$\epsilon = \arcsin\left(\frac{r_D}{a} \sin\varphi_d\right), \tag{4}$$

and

$$r = a \frac{\sin\delta}{\sin\varphi_d}. \tag{5}$$

Using Pythagoras' theorem in the triangle *PAC* allows to calculate *c* as

$$c = \sqrt{(l \cos\alpha - r)^2 + l^2 \sin^2\alpha}, \tag{6}$$

which helps to calculate $\theta$ using the law of cosines:

$$\theta = \arccos\frac{c^2 + l^2 - r^2}{2\,c\,l}. \tag{7}$$

With $\theta$, the torque $T_A$ produced by the spring is easily obtained from Hooke's law applied to torsion springs as

$$T_A = k\,\theta, \tag{8}$$

which gives the force of the spring in the point *P* by division through the lever length *c*:

$$F_P = \frac{T_A}{c}. \tag{9}$$

This needs to be projected into the tangential direction and multiplied with the lever arm to obtain the output torque:

$$T_O = \sin\beta\,\frac{k\,\theta}{c}\,r. \tag{10}$$

The effective stiffness $k_e$ can be defined from this equation as

$$k_e = \frac{\sin\beta\,k\,\theta\,r}{c\,\alpha}, \tag{11}$$

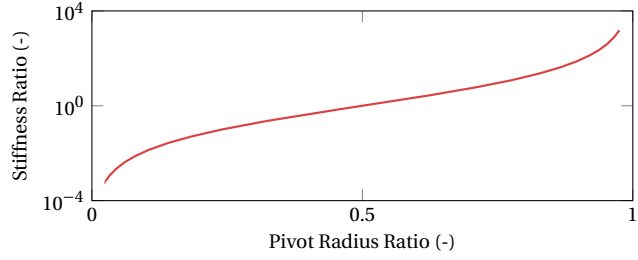

Figure 5: Semi-logarithmic plot for the proposed actuator kinematics of the effective stiffness ratio, defined as effective stiffness $k_e$ divided by spring stiffness $k$, over the pivot radius ratio, defined as pivot radius $r$ divided by lever length $l$. The plot is cropped as a pivot radius ratio of zero yields zero effective stiffness ratio and pivot radius ratio of 1 yields infinite effective stiffness ratio.

such that $T_O = k_e \alpha$.

Further, the motor torques $T_1$ and $T_2$ can be obtained by projection of the forces and the relation between torque and force:

$$F_D = \frac{\cos\beta \, F_P}{\cos\epsilon}, \tag{12}$$

$$F_{T_1} = \sin\beta \, F_P + \text{sign}(\alpha) \, \sin\epsilon \, F_D, \tag{13}$$

and

$$F_{T_2} = \frac{F_D}{\cos\gamma} \tag{14}$$

yield

$$T_1 = (\sin\beta + \text{sign}(\alpha) \, \sin\epsilon \, \frac{\cos\beta}{\cos\epsilon}) \, \frac{k\,\theta}{c} \, r, \tag{15}$$

and

$$T_2 = \frac{\cos\beta \, k \, \theta}{c \, \cos\epsilon \, \cos\gamma} \, r_D. \tag{16}$$

Fig. 5 shows a semi-logarithmic plot of the effective stiffness ratio over the pivot radius ratio, with the stiffness ratio defined as effective stiffness $k_e$ over nominal spring stiffness $k$, and the pivot radius ratio defined as pivot radius $r$ over lever length $l$. For the pivot in the center position (pivot radius ratio of 0.5), the effective stiffness equals the nominal spring stiffness (stiffness ratio of 1). In practice, a pivot radius of zero is not useful to implement with a slider-crank linkage, as it would result in a dead-lock position from which one cannot recover by changing the relative position of stiffness changer and positioning motors. In the final design, the design parameters were set to $r_D = 10\,mm$ $a = 9.5\,mm$, and $l = 20\,mm$, resulting in a theoretical stiffness ratio range from 0.0006 to 1500.

## 2.5. Component Selection and Assembly

Although the kinematics of the actuator are simple, its realization turned out to be rather complex. If the reachable pivot radius should be as small as possible, the pivot axis has to be very close to the shared rotation axis of the motors. Thus, a hollow shaft approach was adopted, meaning that all bearings and driving components were placed on the outside of the actuator, leaving the center region for the pivot mechanism (cf. Figure 2). This was achieved by using large diameter thin section bearings (INA CSEA030) and frameless large diameter thin section motors (ThinGap TG5151). The motors were integrated in direct drive configuration as a concession to not add further complexity by the necessity to also integrate gears.

The final design therefore consists of three disc-shaped modules for the positioning motor, the stiffness changer motor and the damper, which are mounted on a two-parted chassis (cf. figure 7). For position sensing, large diameter magnetic ring absolute position encoders were mounted directly to the respective discs (RLS AksIM). A CAD model of the actuator is provided as Solidworks files in the supplementary material [27].

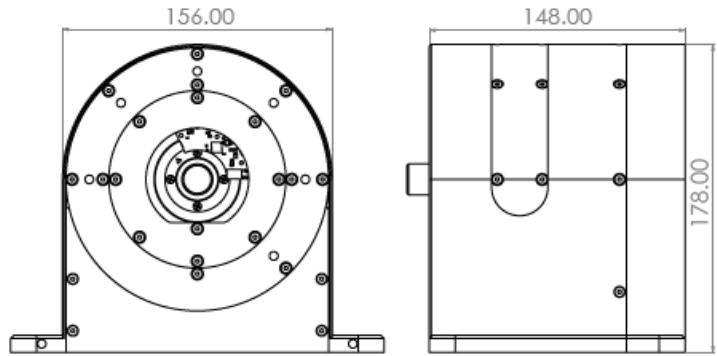

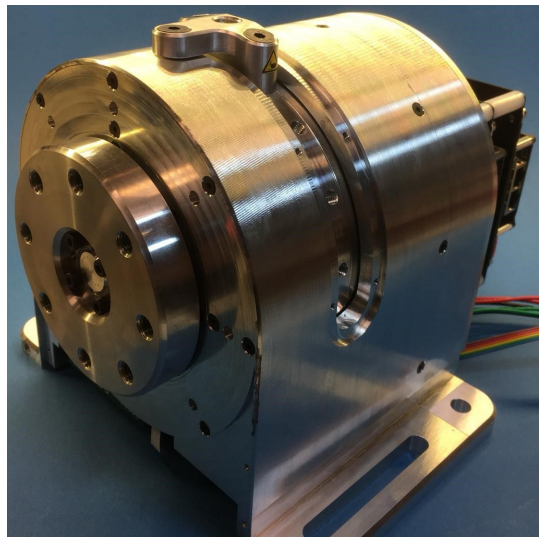

Figure 6: External dimensions and photo of the actuator. The slot in the housing allows to fix the position of motor 2 for calibration purposes with the piece visible on top that can be screwed to the housing and to the motor. The motor drivers are mounted to the back of the housing.

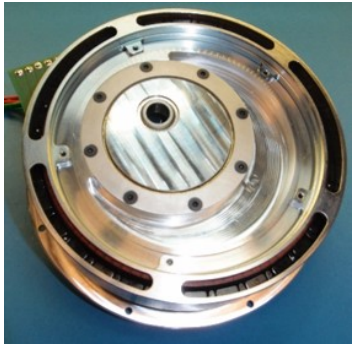

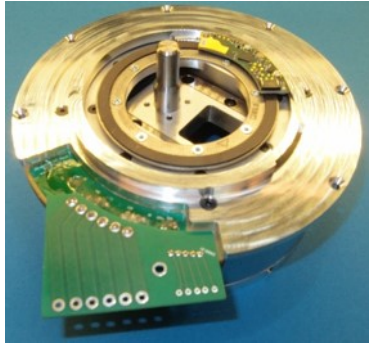

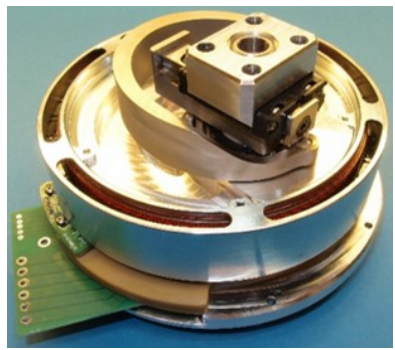

Figure 7: From top to bottom: photos of the prototype stiffness variation module (motor 2 with excentered disc), neutral position variation module (motor 1 with sliding pivot), and elastic module and output module assembled (motor 3 acting as damper, spring and linear guide).

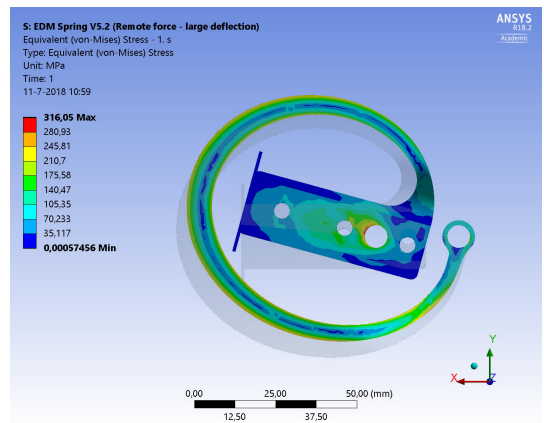

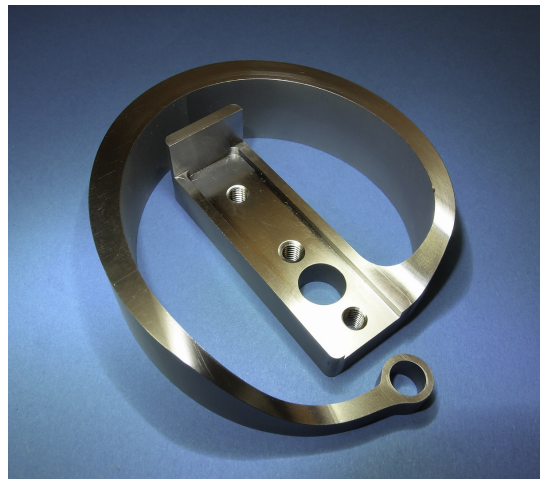

Figure 8: Finite elements simulation result and photo of the final spring design manufactured of titanium grade 5 (Ti6Al4V). The spring shape was iteratively optimized through finite element simulation to meet the desired stiffness of 60 N m/rad while making sure to not exceed the material fatigue strength and not touching surrounding parts at maximum deformation in either direction as the spring is asymmetric.

## 2.6. Spring Design

A monolithic part consisting of a rigid cantilever section and an elastic torsion spring section was designed to act as elastic element, as shown in Fig. 8. A nominal stiffness of $k = 60\,N\,m/rad$ was chosen as design goal for the spring, resulting in a theoretical effective stiffness range of 0.06 N m/rad to 41 000 N m/rad. Considering the selected motors' maximum torque of 12 N m, this corresponds to a maximum spring deflection of 15.1° and a spring torque of 15.8 N m.

Titanium grade 5 (Ti6Al4V) was selected as the spring material because of its capacity to store high potential energy per volume elastically before being affected by plastic deformation. Several torsion spring shapes were considered. Because of spatial constraints, the resulting spring has only one winding. The diameter and thickness of the spring were determined to fit in the design when deformed, and the spring thickness profile was dimensioned as following, such that those values are achieved without exceeding material limits.

First, a finite element simulation was performed in ANSYS Workbench 18 on a spring model with a uniform thickness of 5 mm and a load of 15.8 N m. The simulation result indicated a non-uniform stress distribution over the circumference of the spring. The spring winding thickness was then altered iteratively until an even stress distribution was achieved. By doing so, the stress is uniform along the outer edge of the spring. Second, the spring shape and thickness was optimized further. The height, width and thickness were changed iteratively until the spring had the desired deflection at maximum load (15.1°), the maximum stress did not exceed the titanium grade 5 fatigue strength (450 MPa), and the spring did not touch any surrounding parts at maximum deformation. Since the spring is asymmetric, these conditions were checked in both directions. The spring was manufactured through wire electric discharge machining (EDM) and CNC milling.

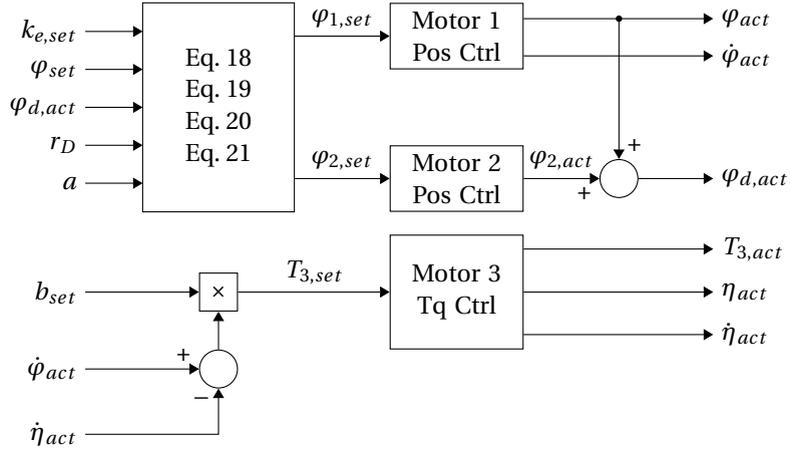

Figure 9: Control diagram of the actuator. The low-level position and torque control are implemented directly in the motor drivers of the respective motors, which receive the set points and return the actual values via EtherCAT to the high-level controller implemented in real-time Linux.

### 2.7. Variable Damper

Motor 3 is acting as a variable damper with the stator mounted to the chassis and the rotor mounted to the output axis. The motor torque $T_3$ is controlled to mimic the behavior of a variable damper with damping factor $b$ between motor 1 and the output shaft:

$$T_3 = b\,(\dot{\varphi} - \dot{\eta})\,. \tag{17}$$

### 2.8. Controller

The motors are controlled by three motor drivers (Ingenia JUP-40/80-E) communicating via EtherCAT with a real-time Linux based high-level controller as shown in figure 9. The EtherCAT loop runs at 1 kHz with the drivers of motor 1 and 2 configured in position control mode and the driver of motor 3 configured in torque control mode. The internal current controllers of the drivers run at 10 kHz.

For the controller, $\varphi_d$ has to be calculated from the radius $r$ to be set for a specific stiffness and the design parameters $a$ and $r_D$ through the law of cosines:

$$\varphi_d = \arccos\left(\frac{a^2 - r_D^2 - r^2}{2\,r_D\,r}\right). \tag{18}$$

Calling $\varphi_1$ and $\varphi_2$ the motor positions of motor 1 and motor 2 respectively, the following equations are also relevant for the controller:

$$\varphi_1 = \varphi\,, \tag{19}$$

and

$$\varphi_2 = \varphi_d - \varphi_1\,. \tag{20}$$

Finally, the controller uses a relationship between $k_e$ and $r$ that was fitted as following to the stiffness measurements of section 3:

$$r = 10^{-3}(0.2273\,(\ln(k_e) + 5.9))^3\,, \tag{21}$$

with $r$ given in m and $k_e$ in N m/rad.

Table 2: Actuator mechanical performance summary

| Measure | Unit | Value |
| --- | --- | --- |
| Nominal Stiffness Variation Time without load | s | 0.09 |
| Nominal Stiffness Variation Time with load | s | 0.12 |
| Nominal Torque | N m | 7 |
| Nominal Velocity | rad/s | 16 |
| Maximum Stiffness | N m/rad | 313 |
| Minimum Stiffness | N m/rad | 0.2 |
| Maximum Damping Torque | N m | 3 |
| Minimum Damping Torque | N m | 0 |
| Maximum Elastic Energy | J | 0.7 |
| Max. Torque Hysteresis | % | 40 |
| Maximum Deflection with max. stiffness | ° | 2.3 |
| Maximum Deflection with min. stiffness (end stop) | ° | 120 |
| Active Rotation Angle | ° | $\infty$ |
| Angular Resolution | ° | $6.9 \times 10^{-4}$ |
| Weight | kg | 10.3 |

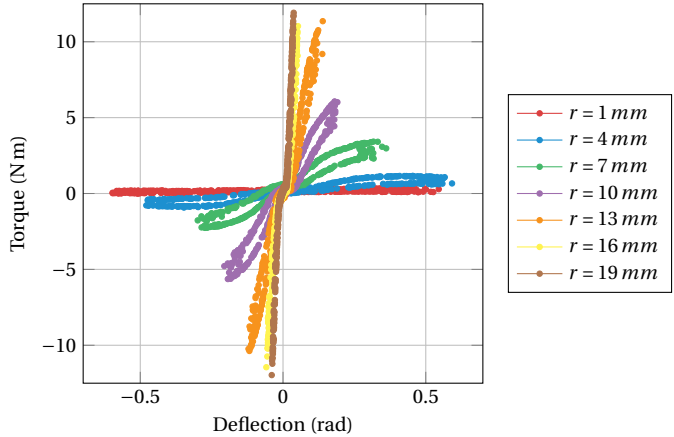

Figure 10: Measured torque over deflection hysteresis plot for different pivot radius settings.

## 3. System Performance

Table 2 shows a summary of the mechanical performance. Figure 10 shows the measured output torque over output deflection plots for different pivot radii, showing the output torque hysteresis. Figure 11 shows the measured effective stiffness over torque plots for different pivot radii with the effective stiffness calculated as output torque over output deflection. Figure 12 shows the measured output torque over output deflection plots for different pivot radii. Figure 10 to 12 use the same measurement data collected with motor 1 fixed and an external torque applied at the output. The torque was measured with an ATI Gamma force-torque sensor mounted to the output axis (calibrated for ± 10 N m torque measurement in z-axis). For figure 11 and 12, the data points were reduced for better visualization by calculating the center line of the hysteresis and applying the Douglas-Peucker line simplification algorithm (Matlab function `reducem`). The dashed lines indicate the theoretical values for reference. Figure 13 shows the measured torque of motor 3 at different velocities for low and high damping settings. Figure 14 shows the measured pivot radius in response to a stiffness set point step, once for a small step and for a large step. The 90 % step response time is 50 ms for a small step and 120 ms for a large step.

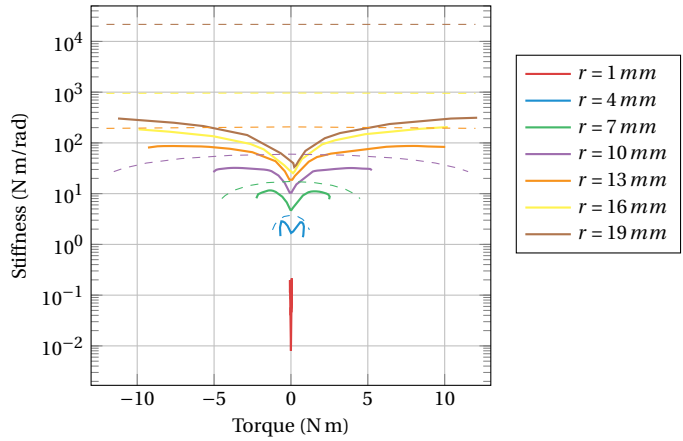

Figure 11: Semi-logarithmic plot of stiffness over external torque. Continuous lines show values extracted from measurement data and dashed lines values as calculated with the design parameters.

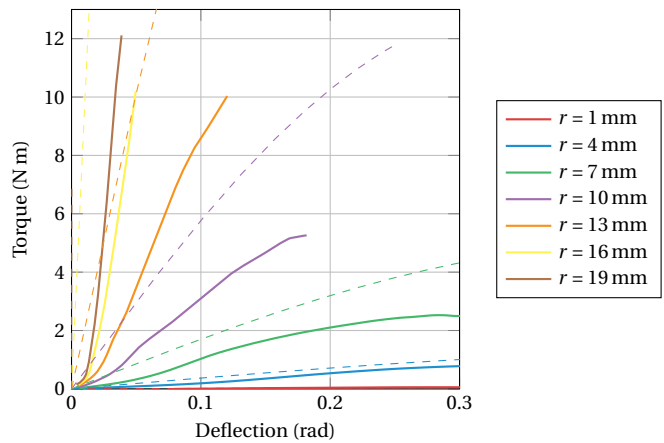

Figure 12: Torque over deflection plot. Continuous lines show values extracted from measurement data and dashed lines values as calculated with the design parameters.

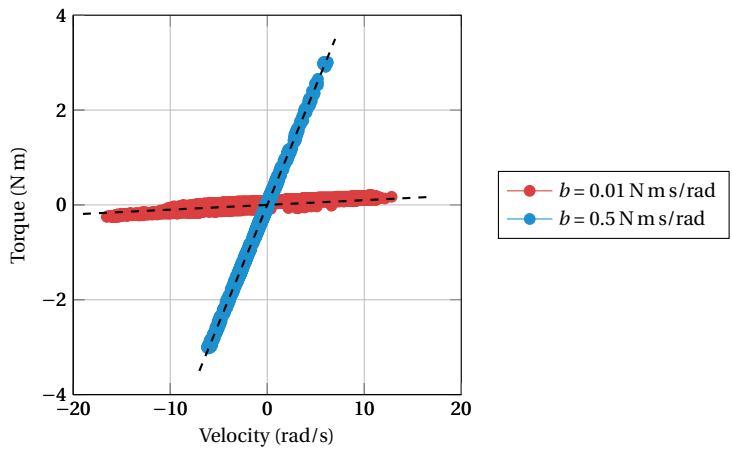

Figure 13: Torque over velocity for low and high damping settings. The dashed lines show the expected torque at the respective damping setting.

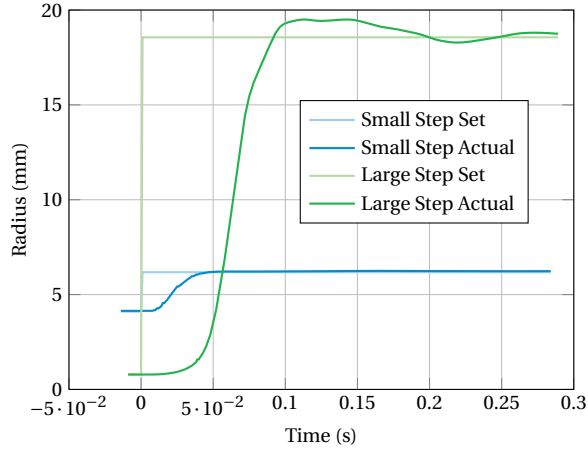

Figure 14: Pivot radius over time for a small and large set point step.

## 4. Application Example

As a simple application example, the actuator was programmed to execute a fast back and forth swinging motion similar to hammering at approximately 3.1 Hz with a cylindrical brass load of 1.6 kg at the output of the actuator. Together with the torque sensor, this resulted in an inertial load of approximately 0.0125 kg m$^2$.

The actuator was programmed to execute the motion once at a high stiffness setting with a pivot radius of 19.1 mm and once with changing the stiffness to a low setting with a pivot radius of 6.9 mm at the beginning of the motion. In both cases, the variable damping setting was set to a low value at the beginning of the motion (0.01 N m s/rad) and triggered to a high value (0.5 N m s/rad) at detection of a negative output position $\eta_{actual}$ to simulate an impact.

Figure 15 shows plots over time of the position, velocity, stiffness and damping torque that were measured during execution of the motion (3.3 s of data were cropped at time 0.5 s, corresponding to a return to initial conditions and waiting for manual start of the next run). At high stiffness ($r_{set}$ = 19.1 $mm$), the deflection is lower than 0.03 rad allowing precise motions, whereas at low stiffness ($r_{set}$ = 6.9 $mm$) the deflection is up to 0.3 rad, helping to reach a higher output velocity than with the stiff setting. The velocity gain obtained with low stiffness is 175 % from input velocity $\dot{\varphi}$ to output velocity $\dot{\eta}$. The stiffness changing time measured was 58 ms for this stiffness change.

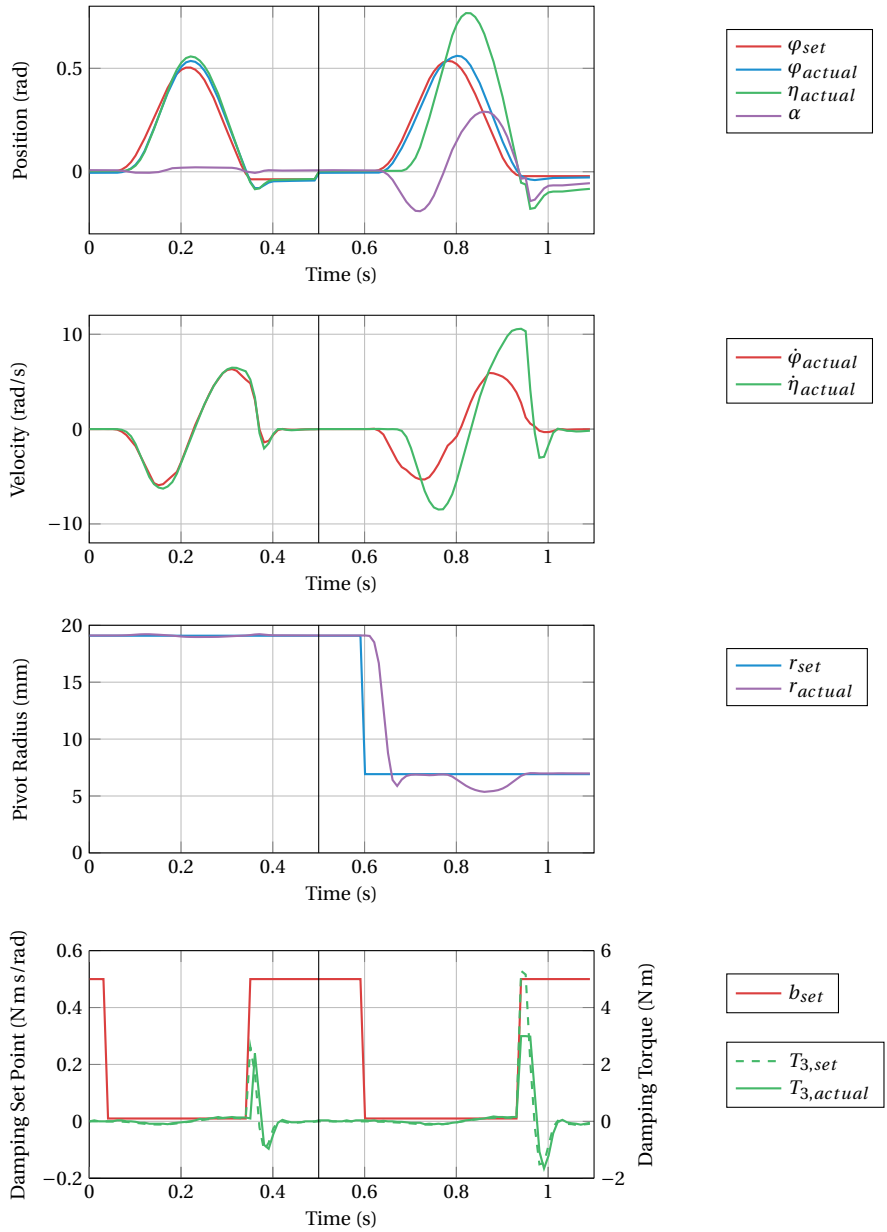

Figure 15: Application example measurements showing a fast back and forth swinging motion similar to hammering at approximately 3.1 Hz, once executed with high stiffness (pivot radius of 19.1 mm) and once with low stiffness (pivot radius of 6.9 mm). The high stiffness setting allows precise motions with low deflection $\alpha$, whereas the low stiffness setting allows to exploit the spring deflection to achieve a higher output velocity $\dot{\eta}$ than the input velocity $\dot{\varphi}$. The fast stiffness changing time of 58 ms shown by the plot of $k_{actual}$ allows to dynamically change the stiffness within a motion. The variable damping is activated at the end of the motion to reduce oscillations, as shown by the plot of the damping set point $b_{set}$ and the corresponding damping torque set point $T_{3,set}$ and measured torque $T_{3,actual}$. Some data were cropped at $t = 0.5\,s$.

## 5. Discussion

In the analysis of whether the design meets the requirements of the tasks, we differentiate between the design and the prototype implementation, which was slightly simplified for budgetary reasons. Most notably, the prototype was implemented without gears for motors 1 and 2, meaning that the position has to be held by the motors in stall operation. This is not an efficient operation mode for electric motors, and negatively affects the positioning performance of the actuator in the presence of external disturbances. However, this was deemed acceptable, as the main mode of operation aimed at with this actuator are highly dynamic tasks. For delicate tasks and highly dynamic tasks, a direct drive configuration is beneficial, as the actuator can then be very soft and suffers less wear than an actuator with gears in unexpected high velocity impacts in a stiff actuator setting, which could occur due to misuse by the operator.

The requirement of 50 ms was only met by the prototype for small stiffness changes, whereas the time measured for large stiffness changes was up to 120 ms. We think that this is not an issue in our targeted bilateral hammering application, because it only requires small stiffness changes as the maximum human arm stiffness was measured to be in the range of 40 N m/rad [28]. The measured damping variation time of 10 ms is due to the motor time constant of the torque controlled motor used for damping.

Also, while the calculated maximum stiffness of the design was 41 000 N m/rad, a maximum stiffness of only 313 N m/rad was measured. This difference is most likely due to play and material deformations in the force transmission. The calculations are based on the assumption that only the spring deforms, while that cannot be guaranteed for this prototype. However, the maximum actuator stiffness is still sufficiently high compared to the maximum human arm stiffness of 40 N m/rad, and therefore the use of Dyrac for bilateral teleoperation is not limited.

Play is an issue of the prototype's performance for explosive movements, as it induces a high torque hysteresis (cf. figure 10), resulting in energy losses in resonance mode. This can explain the lower velocity gain of 175 % in the example application compared to results measured with other VSAs of 272 % [29] and over 400 % [13].

Another effect of the design is that for equal motor torques, the output torque is higher in the negative direction than the positive direction (9.4 N m to 7 N m) because the respective torques of the position changer motor and the stiffness changer motor add up in the negative direction, whereas they subtract in the positive direction. This is a direct consequence of the crankshaft mechanism. While this is in general undesirable, it is in many practical applications not an issue as there is often a working direction requiring a large force and a disengaging direction requiring much less force. Moreover, the configuration can be changed on-the-fly by setting $\varphi_d$ to a negative value to swap the stronger and the weaker direction.

Mechanical end stops were added at ±120° deflection in the prototype to avoid situations that cannot be recovered from by the motors alone. This is mainly due to the crankshaft mechanism leading to very unfavorable configurations for low stiffness settings. Indeed, for a low stiffness setting $\varphi_d$ has to be large, thus a higher torque in motor 2 is required to maintain the equilibrium of forces in Fig. 4 (the cosine terms in the denominator of equation 16 tend to zero for small pivot radii).

The actuator weight was not optimized for this prototype, leading to a heavy actuator (10.3 kg) compared to existing variable stiffness actuators with similar performance, like the FSJ or the AwAS-II (both 1.4 kg). This could be improved in a future design iteration. However, the other actuators also do not have variable damping capability, which adds some mass. Indeed, Dyrac is one of very few implemented actuators that allows full impedance variation (*i.e.*, stiffness and damping) with such high dynamics and over such a wide stiffness range.

The application example shows that the actuator is suitable for precision tasks as well as highly dynamic tasks and that it can dynamically adapt the stiffness as needed. Also the variable damping is very useful to suppress unwanted oscillations without affecting the desired ones, as the whole point of using VIAs for hammering is to exploit the mechanical resonance and for this it is crucial that the damping can be changed to close to zero. At the same time, the goal was to design a versatile robotic actuator and not a specialized hammering tool, thus, oscillations have to be suppressed during precision tasks. The variable damper allows to fulfill both requirements.

Overall the measured actuator performance is very promisiong for performing also human-like dynamic motions such as hammering, shaking, jolting and throwing intuitively and efficiently through teleoperation additionally to delicate tasks and precision tasks. Also, a human-subject study was successfully performed with an accurate simulation model of Dyrac, showing the usefulness of fast stiffness changing time and the variable damping of Dyrac to perform position tasks as well as dynamic tasks through teleoperation efficiently and intuitively in the sense that the human operator does not need to change the mode of operation explicitly [30].

## 6. Conclusion

A new variable impedance actuator design was presented having some attractive performance features for dynamic actuation research:

1. fast impedance changing time of 120 ms for full-range stiffness steps (50 ms for small steps) and 10 ms for full-range damping steps,
2. a wide effective stiffness range of 0.2 N m/rad to 313 N m/rad,
3. variable damping with a damping torque from 0 N m to 3 N m.

A first prototype of the new design was manufactured and characterized to show the feasibility and validating the performance requirements in hardware, albeit with small limitations. The benefits of this actuator for teleoperated applications have successfully been demonstrated in an application example and as a simulation model in a human-subject study [30], showing that the actuator is a versatile variable impedance actuator suitable for both precise positioning tasks and highly dynamic tasks like hammering.

## 7. Acknowledgements

This project was supported by the Dutch Research Council (NWO) under the project grant 12161.